\pdfoutput=1

\documentclass[11pt]{article}

\usepackage[]{acl}

\usepackage{times}
\usepackage{latexsym}

\usepackage[T1]{fontenc}

\usepackage[utf8]{inputenc}

\usepackage{microtype}
\usepackage{tabularx,booktabs}
\usepackage{multirow}
\usepackage{stackengine}
\usepackage{amssymb}

\usepackage{graphicx}
\usepackage{caption}
\usepackage{subfigure}

\usepackage{algorithm}
\usepackage{algpseudocode}
\usepackage{amssymb}
\usepackage{amsmath}
\usepackage{hyperref}

\newcolumntype{P}[1]{>{\centering\arraybackslash}p{#1}}
\newcolumntype{M}[1]{>{\centering\arraybackslash}m{#1}}

\title{In-context Learning Distillation: Transferring Few-shot Learning Ability of Pre-trained Language Models}

\newcommand\tab[1][1cm]{\hspace*{#1}}
\author{Yukun Huang \tab  Yanda Chen \tab Zhou Yu \tab Kathleen McKeown \\
Columbia University \\
\texttt{\{yh3310, zy2461\}@columbia.edu, \{yanda.chen, kathy\}@cs.columbia.edu}
  }

\begin{document}
\maketitle
\begin{abstract}
Given the success with in-context learning of large pre-trained language models, we introduce in-context learning distillation to transfer in-context few-shot learning ability from large models to smaller models.
%
We propose to combine in-context learning objectives with language modeling objectives to distill both the ability to read in-context examples and task knowledge to the smaller models. We perform in-context learning distillation under two different few-shot learning paradigms: Meta In-context Tuning (Meta-ICT) and Multitask In-context Tuning (Multitask-ICT). Multitask-ICT performs better on multitask few-shot learning but also requires more computation than Meta-ICT.
%
Our method shows consistent improvements for both Meta-ICT and Multitask-ICT on two benchmarks: LAMA and CrossFit. 
%
Our extensive experiments and analysis reveal that in-context learning objectives and language modeling objectives are complementary under the Multitask-ICT paradigm. In-context learning objectives achieve the best performance when combined with language modeling objectives. 
\end{abstract}

\section{Introduction}
Large language models have exhibited impressive \textit{in-context learning} ability, where the model performs few-shot learning by conditioning on several input-output pairs (demonstrations) without updating any parameters. 
Despite their remarkable few-shot learning 
performance, large language models always require massive computation resources, which significantly hurts the democratization of NLP for public use.
Large language models like GPT3 are only deployable on extremely large-scale servers for their massive memory usage. 
They also can't be used in real-time systems due to their inefficient inference.

A natural question is whether few-shot learning can be transferred from a large model to a smaller model.
Knowledge Distillation (KD) has been widely proven to be effective in knowledge transfer by teaching a student to mimic a teacher's behavior. 
Zero/Few-shot KD approaches (\citealp{rashid-etal-2021-towards}, \citealp{yoo-etal-2021-gpt3mix-leveraging}) usually leverage the teacher models to augment data, but few of them focus on directly transferring few-shot learning ability. In addition, their student models only specialize in a single task. They have to train and deploy several student models when facing multiple few-shot learning tasks and therefore become inefficient. How well the small model would perform multitask few-shot learning under the supervision of a large language model has not been investigated. 

To answer this question, we propose
in-context learning distillation to transfer the multitask few-shot learning ability.
We distill through both in-context learning objectives and language modeling objectives to transfer knowledge effectively. Specifically, in-context learning distillation helps the student to infer the task based on in-context examples and locate its intrinsic knowledge relevant to the task, while language modeling objectives provide supplementary information about training tasks.  

We investigate the transferability of few-shot learning 
under two different few-shot learning paradigms: Meta In-context Tuning (Meta-ICT) and Multitask In-context Tuning (Multitask-ICT). In Meta-ICT (\citealp{chen-etal-2022-meta}, \citealp{min-etal-2022-metaicl}), the language model is meta-trained on a large collection of tasks through in-context learning objectives and then adapted to unseen target tasks via in-context learning. 
However, in-context learning mostly relies on the knowledge obtained during the pre-training phase \citep{Reynolds2021PromptPF} and doesn't make full use of the input-label correspondence information that is given in the training data \citep{Min2022RethinkingTR}.
To better exploit such information in the few-shot training examples,
we propose another few-shot learning paradigm -- Multitask In-context Tuning. Multitask-ICT first tunes the model through in-context learning objectives with few-shot examples from target tasks and then makes predictions via in-context learning. Multitask-ICT outperforms Meta-ICT but also requires more computation during task adaptations. There is a trade-off between performance and computation for these two few-shot learning paradigms. 

We experiment with in-context learning distillation under 
these two few-shot learning paradigms on two benchmarks: LAMA and CrossFit.  In our experiments, we have 41 different factual and commonsense understanding tasks in LAMA and we have 53 real-life tasks taken from CrossFit including classification, natural language inference, question answering, and so on.
We achieve consistent improvements on both benchmarks compared to in-context tuning without teacher supervision.
For Multitask-ICT, we can reduce the model size by 93\% while retaining the 91.4\% performance of the teacher. Reducing the model size by 68\% or less in fact leads to better performance than the teacher.
%

In summary,
1) We propose In-context Learning Distillation, a teacher-student framework to transfer few-shot learning ability from a large language model to a smaller one. 
2) We propose another new few-shot learning paradigm: multitask in-context tuning, which demonstrates superior performance compared to traditional few-shot supervised fine-tuning and Meta-ICT. 
3) We conduct extensive experiments to understand the role of in-context learning objectives and language modeling objectives from a distillation point of view and find they are complementary to each other.


\begin{figure*}[t]
\centering
\includegraphics[width=1\textwidth]{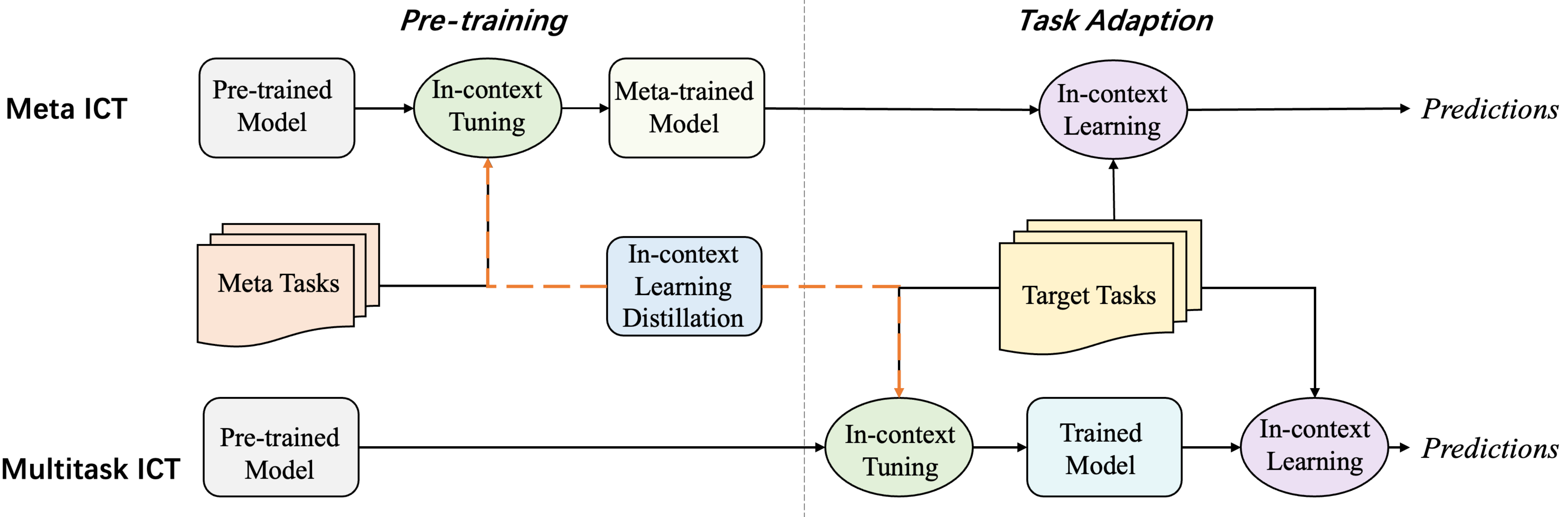}
\caption{In-context tuning paradigms comparison between Meta In-context Tuning and Multitask In-context Tuning. \textit{In-context learning distillation} with dotted lines indicates the distillation process within two paradigms. }
\label{fig: paradigms}
\end{figure*}

\section{Related Work}
\subsection{Knowledge Distillation}
\textbf{KD for Language Models} Knowledge Distillation (KD) was first proposed by \citet{hinton2015distilling} to transfer  knowledge from a teacher model with a high learning capacity to a lower-capacity student model through soft-targets predicted by the teacher. KD has been extensively studied for pre-trained language models. KD can be applied during pre-training (e.g., Distill-BERT, \citealp{Sanh2019DistilBERTAD}), fine-tuning (e.g., BERT-PKD, \citealp{Sun2019PatientKD}), or both (e.g., Tiny-BERT, \citealp{jiao-etal-2020-tinybert}, Distill-GPT2, \citealp{Li2021ASS}).
\\ \\
\textbf{Objectives of KD} Traditional KD optimizes the student model on the objective consisting of the prediction loss over the task labels (hard label) and prediction loss over the teacher's final layer output (soft label). Many efforts have been put into improving the distillation objective to better transfer knowledge. Some work  (\citealp{haidar-etal-2022-rail}, \citealp{Sun2019PatientKD}, \citealp{haidar-etal-2022-cilda}, \citealp{wu-etal-2021-universal}, \citealp{xu-etal-2020-bert}) incorporate intermediate layers matching into the objective function to leverage the knowledge in the hidden layers. Several approaches (\citealp{jafari-etal-2021-annealing}, \citealp{mukherjee-hassan-awadallah-2020-xtremedistil}, \citealp{lu-etal-2021-rw-kd}) adjust the weights of hard label loss and soft label loss in the objective to selectively transfer knowledge. Some methods (\citealp{rezagholizadeh-etal-2022-pro}, \citealp{zhou-etal-2022-bert}) also set objectives for the teacher to better match the student. These improvements are all based on task-specific objectives and we are the first to explore in-context learning objectives in KD. 
\\ \\
\textbf{Zero/Few-shot KD} There are several zero/few-shot knowledge distillation methods for NLP. They (\citealp{rashid-etal-2021-towards}, \citealp{yoo-etal-2021-gpt3mix-leveraging}, \citealp{He2021GenerateAA}) all focus on leveraging teacher models to generate synthetic data when task-specific data is deficient. Instead of using the teacher to generate data, we directly transfer the few-shot learning ability from the teacher to the student. 
Moreover, their student models specialize in one task while our student model is task-agnostic.


\subsection{In-context Learning}
\textbf{Leveraging In-context Learning} In-context learning \citep{Brown2020LanguageMA} performs few-shot learning by doing inference conditioning on a concatenation of input-label examples from the task, without updating any parameters of language models. 
Some follow-up work (\citealp{Zhao2021CalibrateBU}, \citealp{holtzman-etal-2021-surface},\citealp{min-etal-2022-noisy}) further reformulates in-context learning to better perform few-shot learning. 
In addition to few-shot learning, in-context learning can also be leveraged for data augmentation (\citealp{yoo-etal-2021-gpt3mix-leveraging}, \citealp{Chen2022WeaklySD}). 
\\ \\
\textbf{Improving In-context Learning}
In-context learning is found to be over-sensitive and unstable to the choices of in-context examples (\citealp{lu-etal-2022-fantastically}, \citealp{Zhao2021CalibrateBU}, \citealp{Chen2022OnTR}). 
To address this, some work explores methods to find better in-context examples as demonstrations and show decent performance gains (\citealp{rubin-etal-2022-learning}, \citealp{liu-etal-2022-makes}, \citealp{lu-etal-2022-fantastically}). 
Meanwhile, meta in-context tuning (\citealp{chen-etal-2022-meta}, \citealp{min-etal-2022-metaicl}) is proposed to meta-train the model with explicit in-context learning objectives on a wide range of 
tasks to enhance the in-context learning ability and reduce the sensitivity. Our work improves models' in-context learning ability by incorporating supervision from larger models. 
\\ \\

\begin{figure*}[t]
\centering
\includegraphics[width=1\textwidth]{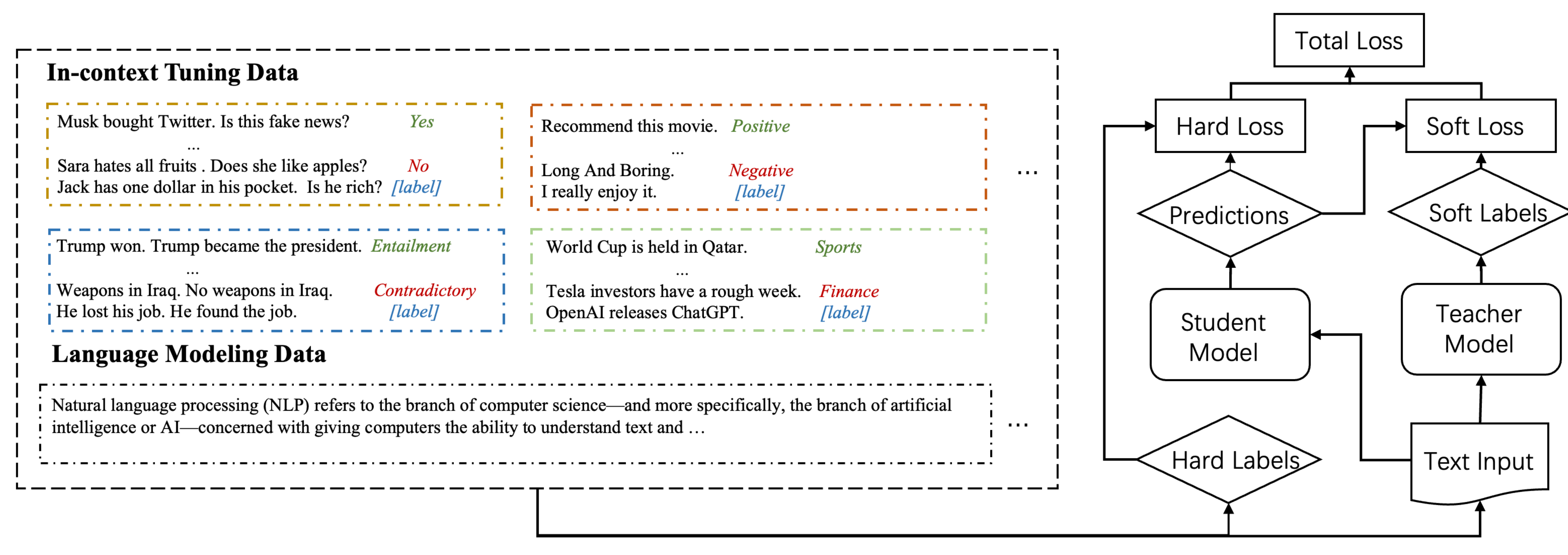}
\caption{Overview of In-context Learning Distillation }
\label{fig: overview}
\end{figure*}

\section{In-context Tuning Paradigms}
In this section, we first describe the background of in-context learning and in-context tuning. Then we introduce two in-context tuning paradigms for few-shot learning: Meta In-context Tuning (Meta-ICT) and Multitask In-context Tuning (Multitask-ICT). 
Meta-ICT is an existing algorithm (\citealp{min-etal-2022-metaicl}, \citealp{chen-etal-2022-meta}) to meta-train LMs on in-context learning objectives to enhance their in-context learning ability. Multitask-ICT is our newly proposed method to adapt LMs to few-shot learning tasks by optimizing in-context learning objectives.
Finally, we compare two paradigms. 

\subsection{Background: In-context Learning/Tuning}
\textbf{In-context learning}
In-context learning refers to the learning procedure where models learn to make predictions for text inputs in a target task by conditioning on a few input-label pairs. Formally, let $k$ be the number of demonstrations, $\{(x_i, y_i)\}_{i=1}^k$ be training samples from a target task and $(x_{k+1}, y_{k+1})$ be the test sample, where $x_i$ is the text and $y_i$ is the label.
The model is given a concatenation of $x_1, y_1, ..., x_k, y_k, x_{k+1}$ as input and predicts $y_{k+1}$. 
\\ \\
\textbf{In-context tuning}
In-context tuning optimizes LMs with in-context learning objectives. Specifically, let $\{(x_i, y_i)\}_{i=1}^{k+1}$ be training samples from a training task $\mathcal{T}$. We feed $x_1, y_1, ..., x_k, y_k, x_{k+1}$ into the model and train the model to generate $y_{k+1}$ using a negative log likelihood objective. 
Formally, assume the number of in-context examples is $k$. For each input text $x$ in a task $\mathcal{T}$, $S_k^{x}$ are the demonstrations consisting of $k$ input-output pairs sampled from the same task $\mathcal{T}$. The in-context learning objective for a task is: 
\begin{align}
\label{equal:ict}
    \mathcal{L}_{\mathcal{T}}^{ICT} &= \sum_{(x,y) \in D_\mathcal{T}} -\log p(y|x, S_{k}^x, \theta)
\end{align}
where $\theta$ is the model parameters. 
\subsection{Background: Meta In-context Tuning}
Meta-ICT trains the model on 
in-context learning objectives with a large collection of tasks. During this process, the model learns to adapt to new tasks through in-context learning. Therefore, we refer to the training process as meta-training and the training tasks as meta-training tasks  $\mathcal{T}_{meta}$. Then the model makes predictions on unseen target tasks $\mathcal{T}_{target}$ via in-context learning. Figure \ref{fig: paradigms} shows the procedure of Meta-ICT.
The total Meta-ICT objective $\mathcal{L}_{meta}^{ICT}$ sums the in-context learning objectives (\ref{equal:ict}) across meta-training tasks:
\begin{align}
    \mathcal{L}_{meta}^{ICT} &= \sum_{\mathcal{T} \in \mathcal{T}_{meta}} \mathcal{L}^{ICT}_{\mathcal{T}}
\end{align} 

Meta-training on the in-context objective has been proven to mitigate over-sensitivity in example ordering, example choices, and instruction wording (\citealp{chen-etal-2022-meta}). 
However, in-context learning during task adaptions ignores part of the label information in the input training data \citep{Min2022RethinkingTR} and therefore doesn't fully make use of the few-shot training examples.


\subsection{Proposed: Multitask In-context Tuning}
To better exploit the information in the few-shot training examples, we propose multitask in-context tuning for multitask few-shot learning.
As shown in Figure \ref{fig: paradigms}, multitask-ICT directly adapts the model to target tasks $\mathcal{T}_{target}$ in two steps. First, it updates model parameters with a few examples from target tasks in an in-context tuning manner. Then it makes predictions via in-context learning. 
The objective for each task is also Equation (\ref{equal:ict}), which is the same as Meta-ICT. The difference is that the total Multitask-ICT objective sums the objective across target tasks instead of meta-training tasks:
\begin{equation}
    \mathcal{L}_{multi}^{ICT} = \sum_{\mathcal{T} \in \mathcal{T}_{target}} \mathcal{L}^{ICT}_{\mathcal{T}}
\end{equation}
where $\mathcal{T}_{target}$ are target few-shot tasks to evaluate.

Due to the limitation on the input sequence length of LMs, a single input sequence sometimes can't fit all the few-shot training samples as in-context examples. In other words, the number of training examples $n$ can be larger than the number of in-context examples $k$. To address this inconsistency, we propose majority voting inference. 
We randomly select $k$ in-context examples from $n$ training samples for $m$ times and choose the most common one from $m$ predictions as the final prediction. We find majority voting can further mitigate the over-sensitivity to the choice of in-context examples and improve the performance. 

\subsection{Paradigm Comparison}
As shown in Figure \ref{fig: paradigms}, 
both Meta-ICT and Multitask-ICT paradigms optimize the in-context learning objectives and perform predictions via in-context learning. However, they differ in why and how they perform in-context tuning.

The goals of in-context tuning for the two paradigms are different. MetaICT performs in-context tuning to prepare the model for adapting to target tasks while Multitask-ICT performs in-context tuning to directly learn the target tasks. 

The two paradigms also differ significantly in their treatment of training and target tasks.
1) the training tasks for Multitask-ICT are the target tasks $\mathcal{T}_{target}$, while the training tasks for Meta-ICT are the meta-training tasks $\mathcal{T}_{meta}$ not overlapping with $\mathcal{T}_{target}$. 2) 
The training examples to update the model for Multitask-ICT are limited,  while for Meta-ICT are abundant.

\section{In-context Learning Distillation}
We propose in-context learning distillation to distill the few-shot learning ability from the teacher to the student. Figure \ref{fig: overview} shows the framework of our approach. We distill through both in-context learning objectives and language modeling objectives. Our approach is compatible with both Meta-ICT and Multitask-ICT paradigms.
As shown in Figure \ref{fig: paradigms}, we perform in-context learning distillation at the in-context tuning stage for both Meta-ICT and Multitask-ICT paradigms.

The student learns from the teacher by imitating the teacher's predictions (\textit{soft labels}). The student learns to perform in-context learning as well as language modeling from the teacher. 
 The soft label loss $\mathcal{L}_{soft}$ measures the discrepancy between the teacher's predictions and the student's predictions, which consists of the in-context learning objective $\mathcal{L}_{soft}^{ICT}$ and the language modeling objective $\mathcal{L}_{soft}^{LM}$
\begin{equation}
    \begin{split}
        \mathcal{L}_{soft} = \mathcal{L}_{soft}^{ICT} + \beta\mathcal{L}_{soft}^{LM}
    \end{split}
\end{equation}
where $\beta$ is the hyper-parameter that balances the in-context learning and language modeling. 
The in-context learning objective is formulated as
\begin{equation}
    \begin{split}
        \mathcal{L}_{soft}^{ICT} = -\sum_{\mathcal{T} \in \mathcal{T}_{train}} \sum_{(x,y) \in \mathcal{D}_\mathcal{T}} \sum_{c \in C_{\mathcal{T}}} \\
    P(y=c|x;S_k^x; \theta^t)\log P(y=c|x;S_k^x; \theta^s) 
    \end{split}
\end{equation}
where $\mathcal{T}_{train}$ represents the training tasks in in-context learning. $\mathcal{T}_{train}$ is $\mathcal{T}_{meta}$ in Meta-ICT and $\mathcal{T}_{target}$ in Multitask-ICT. 
\begin{equation}
    \begin{split}
        \mathcal{L}_{soft}^{LM} = \sum_{x\in \mathcal{D}_{LM}}P(x|\theta^t)\log P(x|\theta^s))
    \end{split}
\end{equation}
where $\mathcal{D}_{LM}$ is a supplemental open domain web text dataset providing general information.

In addition to learning from the teacher's predictions, the student also learns from the ground truths (\textit{hard label}). The hard loss measures the student's performance compared to ground truths, which also consists of both in-context learning objectives and language modeling objectives
\begin{equation}
    \begin{split}
        \mathcal{L}_{hard} = \mathcal{L}_{hard}^{ICT} + \beta\mathcal{L}_{hard}^{LM}
    \end{split}
\end{equation}
\begin{equation}
    \begin{split}
        \mathcal{L}_{hard}^{ICT} = -\sum_{\mathcal{T} \in \mathcal{T}_{train}} \sum_{(x,y) \in \mathcal{D}_\mathcal{T}} 
    \log P(y|x;S_k^x; \theta^s) 
    \end{split}
\end{equation}
\begin{equation}
    \begin{split}
        \mathcal{L}_{hard}^{LM} = -\sum_{x\in \mathcal{D}_{LM}}\log P(x|\theta^s))
    \end{split}
\end{equation}

The final objective function for in-context learning distillation can be formulated as:
\begin{align}
    \mathcal{L}_{KD} = \alpha(t)\mathcal{L}_{hard}+ (1-\alpha(t))\mathcal{L}_{soft} 
\end{align}
We linearly decrease the weight of hard-label loss $\alpha(t)$ and linearly increase the weight of soft-label loss during training.

\section{Experiments}

\subsection{Datasets and metrics}
We utilize two different collections of datasets in this work. The first collection is LAMA \citep{petroni-etal-2019-language}, consisting of 41 tasks for factual and commonsense knowledge understanding. The second collection of datasets includes real-life tasks taken from CrossFit \citep{ye-etal-2021-crossfit}, a few-shot open gym consisting of 160 diverse few-shot NLP tasks. We take 53 unique tasks including text classification, question answering, natural language inference, and paraphrase detection. For LAMA, we adopt mean precision at one and mean precision at ten as our evaluation metrics and report the average scores across tasks. For CrossFit, we adopt Macro-F1 and Accuracy as evaluation metrics for classification tasks and non-classification tasks respectively. 

In addition to the above datasets for in-context learning distillation, we also have auxiliary datasets for language modeling to provide supplemental knowledge. We leverage WikiText \citep{merity2016pointer} as the auxiliary dataset for LAMA and OpenWebText \citep{Gokaslan2019OpenWeb} for CrossFit. 

See Appendix \ref{sec:appendix for datasets} for  details of the datasets.

\subsection{Few-shot Learning Settings}
We experiment with two paradigms (Meta-ICT, Multi-ICT) on two benchmarks (LAMA, CrossFit), resulting in four different few-shot learning settings in total. 
\\
\textbf{Setting 1: Meta-ICT on LAMA} 
We randomly partition 41 tasks into 30 training tasks, 5 validation tasks, and 6 test tasks. We meta-train the model with 30 meta-train tasks and test on 6 target tasks.   
\\
\textbf{Setting 2: Meta-ICT on CrossFit}
We follow the classification to classification setting in \citet{min-etal-2022-metaicl}, where 43 meta-training tasks and 20 target tasks are all classification tasks.  But there are some target tasks where both teacher's and the student's performances are close to random guesses. Therefore, we select 11 out of 20 tasks on which both teacher and student perform better than random guesses as our target tasks. 
\\
\textbf{Setting 3: Multitask-ICT on LAMA} We utilize 41 LAMA tasks as target tasks. We randomly select 60\% of the data for test. Then we randomly select 32 training samples and 32 validation samples from the rest of 40\% of the data. 
\\
\textbf{Setting 4: Multitask-ICT on CrossFit}
We select 18 different tasks as our target tasks, which cover classification, question answering, natural language inference, and paraphrasing.  

Table~\ref{few-shot setting comparison} shows detailed comparisons among four settings. For Setting 2, we follow the \citet{min-etal-2022-metaicl} to utilize channel in-context learning, where the order of the input text and the label is reversed. See Appendix ~\ref{sec:appendix for settings} for setting details and Appendix~\ref{sec:appendix for channel} for more information about channel in-context learning.

\begin{table}[t]
\small
\begin{tabularx}{1.0\linewidth}{XXXXX}
\toprule
Setting & 1 & 2 & 3 & 4 \\
\midrule
Data & LAMA & CROSSFIT & LAMA & CROSSFIT \\
Arch & BERT & GPT2 & BERT & GPT2 \\
Paradigm & Meta & Meta & Multitask & Multitask \\
\#Tasks$_{train}$ &30 &43 &41 &18 \\
\#Tasks$_{test}$ & 6 &11 &41 &18 \\
Overlap & $\times$ & $\times$ & $\checkmark$ & $\checkmark$ \\
k & 5 & 4 & 5 & 4 \\
\#Fewshot & 5 & 4 & 32 & 32 \\
Order & direct & channel & direct & direct \\
\bottomrule
\end{tabularx}
\caption{\label{few-shot setting comparison}
Comparison among four different few-shot settings. Overlap: If the training tasks and target tasks overlap. k: the number of in-context examples in each input to the model. \#Few-shot: the number of available training samples for each task. Order: the order of the text x and the label y for in-context learning formulation. 
}
\end{table}

\subsection{Experiment Details}
All implementation is done with PyTorch and Transformers. To show our method is effective in different architectures, we use BERT as backbone models for experiments on LAMA datasets (BERT-small [25M parameters], BERT-base[110M], BERT-large[336M]) and GPT2 for experiments on Crossfit datasets (GPT2-small [124M], GPT2-medium [355M], GPT2-large [774M]. The hyperparameters and training details can be found in Appendix \ref{sec:appendix for training details}. 

\section{Results}
We first compare three few-shot learning paradigms in Section 6.1. Then we present results for in-context learning distillation in Section 6.2. Finally, we discuss the ablation study in Section 6.3.
\begin{table}[t]
\small
\begin{tabularx}{1.0\linewidth}{X|XXX}
\toprule
Model & Multitask$_{FT}$ & Meta$_{ICT}$ & Multitask$_{ICT}$ \\
\midrule
GPT2$_{lg}$ & 59.8 & 58.7 & \textbf{63.9} \\
GPT2$_{md}$ & 59.7 & 53.6 & \textbf{62.6} \\
GPT2$_{sm}$ & 56.5 & 52.9 & \textbf{57.3} \\
\bottomrule
\end{tabularx}
\caption{\label{tab: few-shot learning paradigm comparison}
Results of three different few-shot learning paradigms. Multitask$_{FT}$: multitask fine-tuning with few-shot examples. Meta$_{ICT}$: meta in-context tuning.  Multitask$_{ICT}$: multitask in-context tuning. Experiments are conducted on setting 2. We randomly select 32 training samples and 32 validation samples from each of the nine target tasks. We report the average F1 scores and mark the best performance as \textbf{bold}. 
}
\end{table}

\begin{table*}[t]
\centering
\small
\begin{tabularx}{1.0\textwidth}{X|XX|XXXX}
\toprule
\textbf{Data}& \textbf{Teacher} & \textbf{ICT} &\textbf{Student} &\textbf{ICT(Baseline)}  & \textbf{ICL-D} & \textbf{ICL-D(LM)} \\
\midrule
\multirow{4.5}{*}{\shortstack[l]{Setting 1 \\ LAMA}}
&\multirow{2}{*}{BERT-large } &\multirow{2}{*}{28.3/58.7} & BERT-base & 23.9/53.8 & \textbf{25.2/56.2}  & 24.3/55.4   \\[1.25ex]
& & & BERT-small & 14.9/47.7 & \textbf{16.8/49.1}  & 15.6/48.8  \\ [1.25ex]
\cmidrule{2-7}
&\multirow{2}{*}{BERT-base} & \multirow{2}{*}{23.9/53.8} & \multirow{2}{*}{BERT-small} & \multirow{2}{*}{14.9/47.7} & \multirow{2}{*}{\textbf{16.2/49.2}}  & \multirow{2}{*}{15.4/48.8}  \\ 
& & & & & & \\
\midrule [0.75pt]
\multirow{4.5}{*}{\shortstack[l]{Setting 2 \\ CrossFit}}
&\multirow{2}{*}{GPT2-large } &\multirow{2}{*}{57.0} & GPT2-med & 51.4 & \textbf{53.6}  & 53.6   \\[1.25ex]
& & & GPT2-small & 50.2 & \textbf{51.3}  & 51.1  \\ [1.25ex]
\cmidrule{2-7}
&\multirow{2}{*}{GPT2-med} & \multirow{2}{*}{51.4} & \multirow{2}{*}{GTP2-small} & \multirow{2}{*}{50.2} & \multirow{2}{*}{50.8}  & \multirow{2}{*}{\textbf{52.1}}  \\ 
& & & & & & \\
\bottomrule
\end{tabularx}

\caption{\label{tab: meta-ict}
Distillation results under Meta-ICT paradigms. ICT(baseline): In-context tuning without the teacher. ICL-D: In-context knowledge distillation which only distills through in-context learning objectives. ICL-D(LM): In-context knowledge distillation which distills through both in-context learning objectives and language modeling objectives. Two numbers on LAMA datasets indicate the average precision@1 and precision@10 
of all target tasks. The number on CrossFit indicates the average macro F1 score of all target tasks. The number of in-context examples for setting 1 is k=5, while for setting 2 is k=4, therefore setting 1 is 5-shot, and setting 2 is 4-shot. \textbf{Bold} indicates the best score.
}
\end{table*}

\begin{table*}[t]
\centering
\small
\begin{tabularx}{1.0\textwidth}{X|XX|XXXX}
\toprule
\textbf{Data}& \textbf{Teacher} & \textbf{ICT} &\textbf{Student} &\textbf{ICT(Baseline)}  & \textbf{ICL-D} & \textbf{ICL-D(LM)} \\
\midrule
\multirow{4.5}{*}{\shortstack[l]{Setting 3 \\ LAMA}}
&\multirow{2}{*}{BERT-large } &\multirow{2}{*}{25.6/57.3} & BERT-base & 22.3/53.2 & 23.7/55.4 & \textbf{27.2/61.2}  \\[1.25ex]
& & & BERT-small & 12.1/43.0 & 13.6/45.8  & \textbf{18.5/52.4}  \\ [1.25ex]
\cmidrule{2-7}
&\multirow{2}{*}{BERT-base} & \multirow{2}{*}{22.3/53.2} & \multirow{2}{*}{BERT-small} & \multirow{2}{*}{12.1/43.0} & \multirow{2}{*}{13.7/45.8}  & \multirow{2}{*}{\textbf{18.5/52.3}}  \\ 
& & & & & & \\
\midrule [0.75pt]
\multirow{4.5}{*}{\shortstack[l]{Setting 4 \\ CrossFit}}
&\multirow{2}{*}{GPT2-large } &\multirow{2}{*}{66.2} & GPT2-med & 64.3 & 65.5  & \textbf{65.8}   \\[1.25ex]
& & & GPT2-small &58.4 & 59.9  & \textbf{61.2}  \\ [1.25ex]
\cmidrule{2-7}
&\multirow{2}{*}{GPT2-med} & \multirow{2}{*}{64.3} & \multirow{2}{*}{GTP2-small} & \multirow{2}{*}{58.4} & \multirow{2}{*}{59.0}  & \multirow{2}{*}{\textbf{60.5}}  \\ 
& & & & & & \\

\bottomrule

\end{tabularx}

\caption{\label{tab: multitask-ict}
Distillation results under the Multitask-ICT paradigm. Definitions of ICT, ICL-D and ICL-D(LM) are the same as Table \ref{tab: meta-ict}.
We randomly select 32 training samples and 32 validation samples. The evaluation metric and abbreviation is the same as Table \ref{tab: meta-ict}. Though the number of in-context examples is k=5 for LAMA and k=4 for CrossFit, we have access to 32 training samples for each task. Hence both settings are 32-shot learning. 
}
\end{table*}

\subsection{Few-shot learning paradigm comparison}
We compare three few-shot learning paradigms: Multitask-ICT, Meta-ICT, and Multitask supervised-finetuning (Multitask-FT).
\\ \\
\textbf{Multitask-ICT outperforms Meta-ICT in accuracy}
As shown in Table \ref{tab: few-shot learning paradigm comparison}, Multitask-ICT outperforms Meta-ICT across all three scales. The score difference between Multitask-ICT and Meta-ICT can even be as large as 9\% on the GPT2-medium. Multitask-ICT adapts the model to target tasks by tuning the model parameters with a few training samples, while Meta-ICT adapts the model only via inference. The significant advantage of the Multitask-ICT over Meta-ICT indicates that it's still better to update the model even if the model has limited access to the target training samples.
\\ \\
\textbf{Multitask-ICT outperforms Multitask-FT with few-shot training samples}
From Table~\ref{tab: few-shot learning paradigm comparison}, Multitask$_{ICT}$ is better than Multitask$_{FT}$ across all three scales, which suggests that tuning the in-context learning objectives can better exploit limited training examples. We suspect that tuning the in-context objectives encourages the model to utilize the relationships among the datapoints in the same task. In Multitask-ICT, different datapoints from the same task are concatenated 
in a single input to the model. Therefore, the model can capture the relations between datapoints from the same task. On the contrary, in traditional multitask fine-tuning, the model only learns one datapoint at each time and overlooks the relations between these datapoints. 


\subsection{Distillation Results}
\textbf{In-context learning distillation improves in-context learning} 
As shown in Table~\ref{tab: meta-ict} and Table~\ref{tab: multitask-ict}, in-context learning distillation  consistently improves 
both in-context tuning paradigms on both LAMA and CrossFit datasets. ICL-D outperforms ICT in all the settings, indicating that the teacher indeed can transfer helpful  knowledge through in-context learning objectives.
\\ \\
\textbf{Stronger students gain relatively more}
The student with larger capacities gets relatively more benefits from the teacher under the Meta-ICT paradigm. Considering the teacher-student performance gap, the medium-size model(BERT-base, GPT2-medium) gains relatively more than the small-size model (BERT-small, GPT2-small) when taught by a large-size teacher. 
In Setting 2 of Table \ref{tab: meta-ict}, GPT2-medium improves +2.2/5.6 (where 5.6 is the teacher-student gap) while GPT2-small only improves +1.1/6.8. In Setting 1 of Table \ref{tab: meta-ict},  BERT-base improves +2.5/4.9 at precision@10, while BERT-small improves +1.4/11.0.
We hypothesize that larger models have a larger room for in-context learning ability improvement since strong in-context learning ability usually emerges in large models \citep{shin-etal-2022-effect}.
\\ \\
\textbf{Language modeling distillation helps Multitask-ICT} 
Incorporating language modeling distillation brings compelling improvement to the Multitask-ICT paradigm. 
As shown in Table \ref{tab: multitask-ict}, the improvement of ICL-D(LM) over ICT in LAMA datasets is especially prominent: +4.9/+8.0 for BERT-base and +6.4/9.4 for BERT-small. ICL-D(LM) also shows effectiveness on CrossFit datasets. Improvements of ICL-D(LM) over ICT for GPT2-medium (+1.5) and GPT2-small (+2.8) are both larger than improvements of ICL-D over ICT.
In contrast, under the Meta-ICT paradigm, rather than improving results, language modeling distillation even hurts performance to some extent. As shown in Table~\ref{tab: meta-ict}, the ICL-D(LM) gets comparable or even worse results than ICL-D in 5 out of 6 teacher-student pairs.  
We conjecture that language modeling distillation only contributes to the tasks seen during training, 
and meta-training tasks and target tasks don't overlap in Meta-ICT. Learning too much content of meta-training tasks would hinder the generalization of the target tasks.
\\ \\
\textbf{Stronger Teachers don't necessarily lead to better students}
When the teacher changes from the medium size to the large size, the difference between the small size students' performances is negligible. 
Take Setting 2 in Table \ref{tab: meta-ict} as an example, when we increase the teacher from GPT2-medium to gpt2-large (parameter number from 355M to 774M, teacher accuracy from 51.4 to 57.0), the change of GPT2-small's performance is within one point. We observe similar phenomena in all other three settings. This observation is in line with 
other knowledge distillation research (\citealp{Sun2019PatientKD}, \citealp{Yuan2019RevisitKD}). Presumably, the knowledge from the large teacher might be too "hard" for the student to learn and therefore the student can't get further improvement. 
\\ \\
\subsection{Ablation Study}
\begin{figure}[t]
    \centering
    \subfigure[\#training samples]{ \label{fig: ablation1}
        \includegraphics[width=0.48\linewidth]{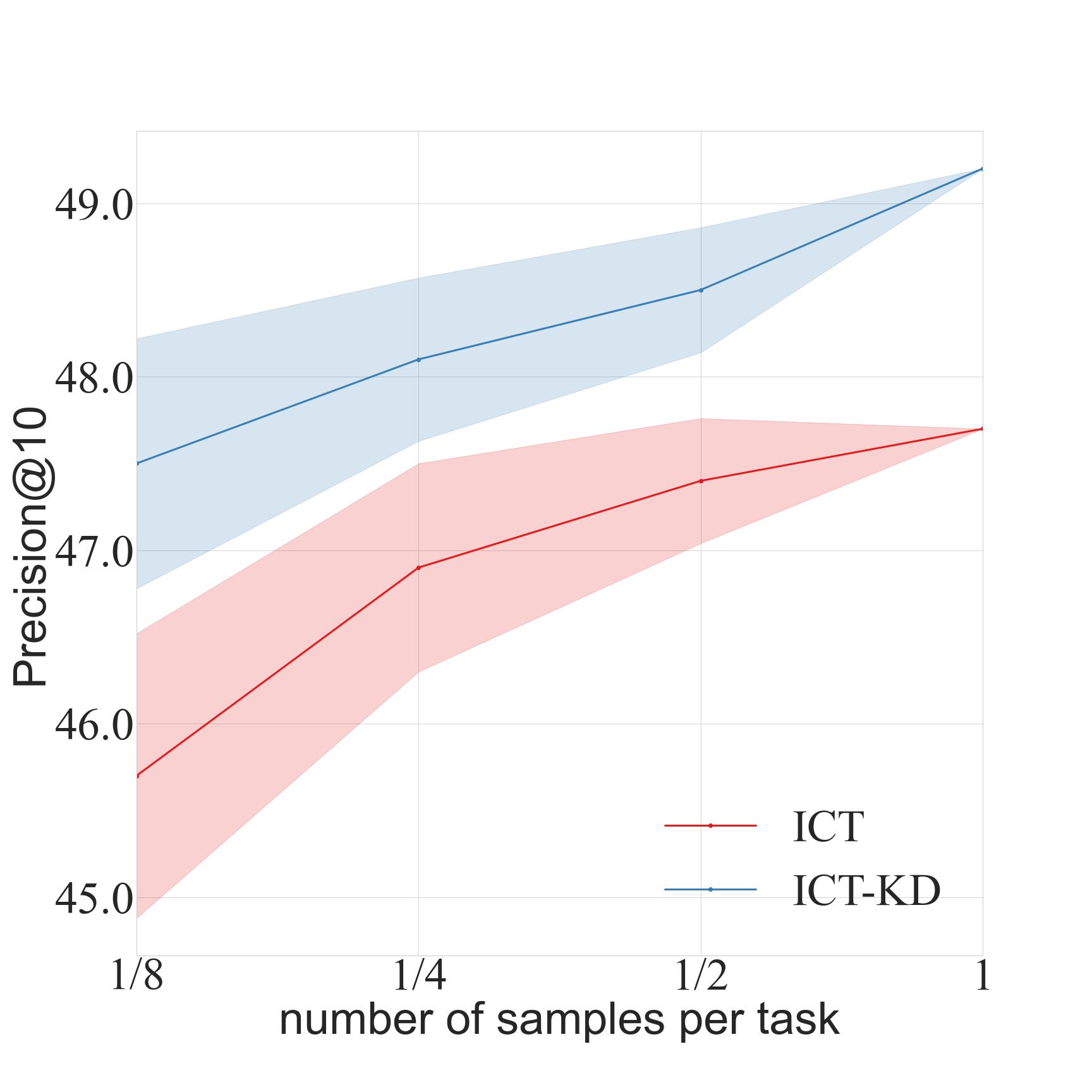}}
    \subfigure[\#meta-training tasks]{ \label{fig: ablation2}
	\includegraphics[width=0.48\linewidth]{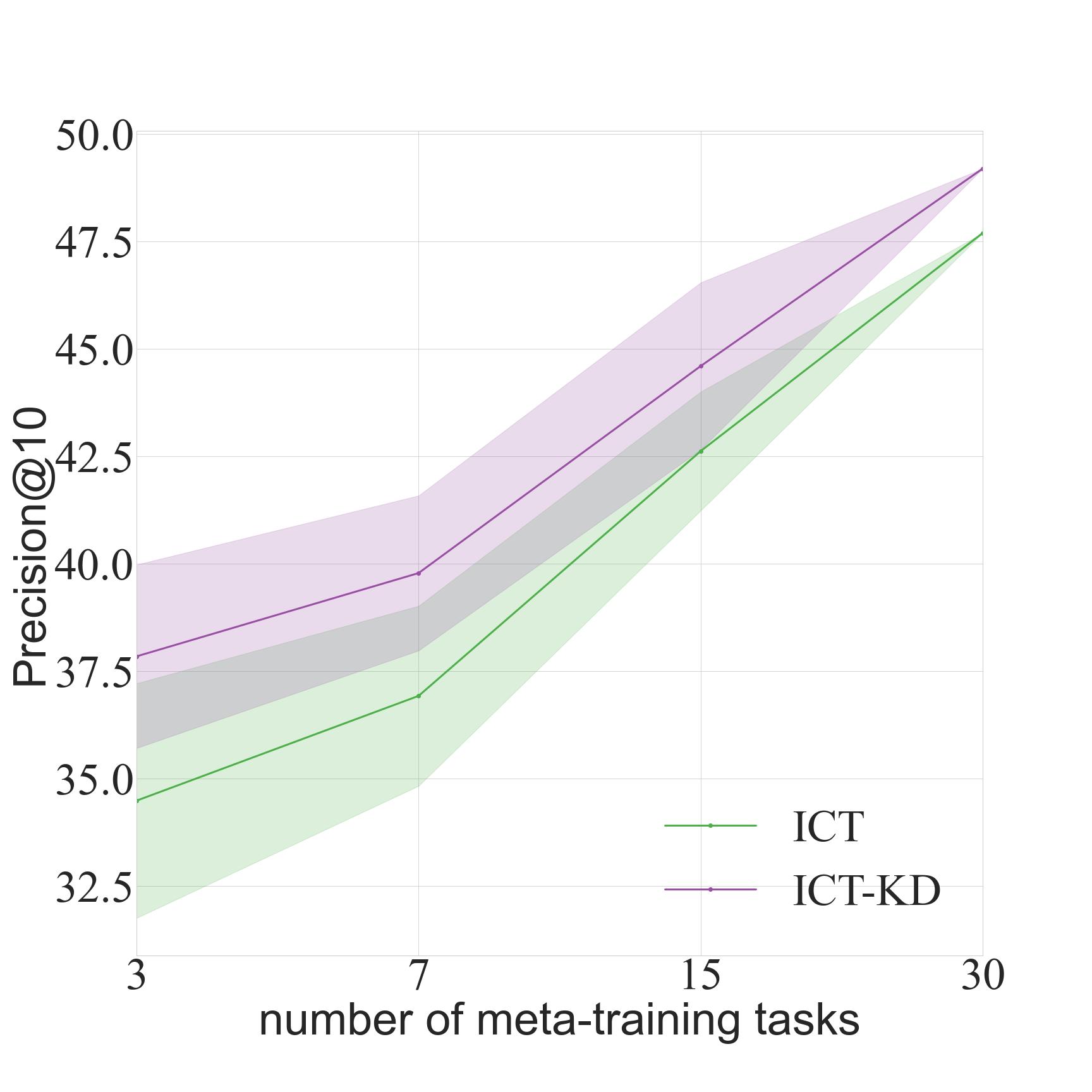}}
    \caption{(a) Meta-ICT results on 12.5\%, 25\%, 50\% and 100\% training datapoints sub-sample levels. The experiment is conducted on Setting 1. We run experiments with five random seeds and report average scores along with standard deviation (b) Results of Meta-ICT trained with $n$ meta-training tasks, where $n$ is in [3, 7, 15, 30]. For each $n$, we sample $n$ meta-training tasks five times with different random seeds and report average scores along with standard deviation. The green line denotes in-context tuning while the purple line represents in-context learning distillation}
    \label{fig: ablation}
\end{figure}

We explore how Meta-ICT generalizes to unseen tasks by examining the influence of the number of datapoints and the number of tasks for meta-training tasks.  Then, we investigate why language modeling objectives help Multitask-ICT by distilling only through language modeling objectives. 
\\ \\
\textbf{Number of training samples}
We investigate how the number of samples for each meta-training task affects our method. We sub-sample [1/8, 1/4, 1/2, 1] of the training samples from each meta-training task to train the model (See Appendix \ref{sec:appendix for datasets} for the number of total training samples for each task). 
According to Figure \ref{fig: ablation1}, ICL-D consistently outperforms ICT when the number of training samples varies. Surprisingly, the performance of Meta-ICT doesn't decrease much when we only utilize 1/8 of the training samples to train the model. This finding suggests that Meta-ICT doesn't rely much on the contents of each meta-training task. We suspect that the model is more likely to learn to read the in-context format instead of learning the contents of meta-tasks during meta-training.
\\ \\
\textbf{Number of meta-training tasks}
To further explore how the number of meta-training tasks influences performance, we vary the number of meta-train tasks from 3 to 30 in the LAMA settings. We run our experiments with five different random sets of meta-training tasks because the Meta-ICT is sensitive to the distributions of selected meta-training tasks. As shown in Figure \ref{fig: ablation2}, our method consistently improves the baseline no matter how many meta-training tasks we include. ICL-D improves more when the number of meta-training tasks is smaller, which suggests that teacher supervision can make up for the deficiency of meta-training tasks to some extent. It should be noted that in-context tuning relies more on the number of tasks than the number of data points for each task. 
When we use only 1/8 of tasks, the performance drops more than 10\%, while the performance only drops around 2\% when we use 1/8 of the data points for each task. This result indicates that the model learned mainly from the distributions of tasks instead of the exact contents of each task. 
\\ \\
\textbf{Language-modeling-only distillation}
Language modeling distillation brings significant improvement to Multitask-ICT. To understand language modeling's role in our approach, we run experiments with only language modeling distillation without in-context learning objectives on Setting 3. From Table \ref{tab: language only distillation}, language modeling distillation has a similar performance to the raw pre-trained language model without any training, which lags far behind other in-context tuning methods. This observation is in line with the finding that language modeling ability doesn't always correlate with in-context learning \citep{shin-etal-2022-effect}. This result suggests that the student can't extract useful knowledge for target tasks from the rich but ill-assorted information provided by the teacher only through language modeling objectives. Language modeling distillation only helps when combined with in-context learning distillation. 

Our results imply that in-context learning objectives and language modeling objectives are complementary to each other during knowledge distillation under the Multitask-ICT paradigm. 
On the one hand, in-context learning distillation guide language modeling distillation. The in-context learning objectives allow the model to learn to locate the relevant task knowledge given the in-context examples. Therefore, the model can locate and absorb useful knowledge from the rich but ill-assorted information contained in language modeling data and the teacher. 
On the other hand, language modeling distillation enriches the in-context learning distillation. The in-context examples for target tasks are deficient under few-shot settings, thereby limiting the improvement brought by in-context learning distillation. Language modeling objectives serve as a carrier to transfer rich information from the teacher to the student. 
\begin{table}[t]
\centering
\small
\begin{tabularx}{0.6\linewidth}{X|X}
\toprule
Method & P@1/P@10 \\
\midrule
Raw & \ \ 1.7/21.5 \\
LM-KD & \ \ 1.6/21.5\\
ICT & 12.1/43.0 \\ 
ICL-D & 13.7/45.8\\
ICL-D(LM) & \textbf{18.5/52.4}\\
\bottomrule
\end{tabularx}
\caption{\label{tab: language only distillation}
Comparison of language modeling distillation to other in-context tuning methods on Setting 3. Raw: the pre-trained language model without any fine-tuning. LM-KD: distill only through language modeling objectives without in-context learning objectives and then perform in-context learning. ICT: Multitask in-context tuning. ICL-D: in-context learning distillation. ICL-D(LM): in-context learning distillation with language modeling distillation. \textbf{Bold} indicates the best score.
}
\end{table}

\section{Conclusion}


In this paper, we present 
an in-context learning distillation framework to transfer the few-shot learning ability from the large language model to the smaller one. Specifically, we distill knowledge through both in-context learning objectives and language modeling objectives. We apply this method to two few-shot learning paradigms: Meta-ICT and Multitask-ICT. Multitask-ICT is a few-shot learning paradigm proposed by us to further improve Meta-ICT but at the cost of computation. In-context learning distillation consistently improves both LAMA and CrossFit benchmarks under two few-shot learning paradigms. For Multitask-ICT, one of our models achieves better performance while being 3.4$\times$ smaller and 1.7$\times$ faster than the teacher, and another one retains 91.4\% performance while being 13.4$\times$ smaller and 3.6$\times$ faster than the teacher. In the ablation study, we find that 1) Meta-ICT learns more from the distributions of meta-training tasks than the contents of each meta-training task.
2) In-context learning distillation performs best when combined with language modeling distillation under the Multitask-ICT paradigm. 

\bibliography{anthology,custom}
\bibliographystyle{acl_natbib}

\appendix

\section{Training Details}
\label{sec:appendix for training details}
Our implementations are based on PyTorch. We also refer to other open-source codes (\citealp{min-etal-2022-metaicl}, \citealp{chen-etal-2022-meta}, \citealp{yang-etal-2020-textbrewer}).  To show our method is effective in different architectures, we use BERT as backbone models for experiments on LAMA datasets (BERT-small [25M parameters], BERT-base[110M], BERT-large[336M]) and GPT2 for experiments on Crossfit datasets (GPT2-small [124M], GPT2-medium [355M], GPT2-large [774M]. We provide training details for all four settings. 
\\
\textbf{Meta-ICT on LAMA}: Our hyperparameters for Meta-ICT and Multitask-ICT are the same. For all methods (ICT, ICL-D, ICL-D(LM)), we choose adamw optimizer with the learning rate 3e-6. We have a linear scheduler for the learning rate with 100 warmup steps.
For ICT and ICL-D, we set epochs to 60, and batch size to 48. We evaluate validation sets every epoch and use the precision@10 on the validation set as an early stopping indicator. We set patience to 2. For ICL-D(LM), we set the max steps to 30000 and evaluate every 5000 steps. We keep the batch size to 48 and set gradient accumulation steps to 5 so that model can see more language modeling data for each update. We combine the language modeling distillation with in-context learning distillation by randomly sampling from in-context learning data and language modeling model data in the weight of [0.1, 0.9] at each step.  
For ICL-D and ICL-D(LM), we set the temperature for knowledge distillation to 2. We linearly grow hard label weight from 0 to 1 and linear decay soft label weight from 1 to 0. \\
\textbf{Meta-ICT on CrossFit} We apply adamw optimizer with the learning rate 1e-5. We set the total batch (batch size $\times$ gradient accumulation steps) to 16 for all methods across all scales. But the exact number of batch size and gradient accumulation steps varies across different scales due to the memory limitation (GPT2-small is 8$\times$2, GPT2-medium is 4$\times$4, and GPT2-large is 2$\times$8). For ICT and ICL-D, We set the max steps to 5000. For ICL-D(LM), we set the max steps to 30000 and sample the in-context learning data and language modeling in the weight of [0.1, 0.9] at each step. For ICL-D and ICL-D(LM), we set the temperature for knowledge distillation to 2. We linearly grow hard label weight from 0 to 1 and linear decay soft label weight from 1 to 0.\\
\textbf{Multitask-ICT on LAMA} We apply the same parameters as Meta-ICT on LAMA. We randomly select 32 validation samples from each target tasks as our validation set. \\  
\textbf{Multitask-ICT on CrossFit} Most parameters are the same as Meta-ICT on CrossFit. We randomly select 32 validation samples from each target task as our validation set. We set the patience to be 2. For ICT and ICL-D, we set the max steps to 2000 and evaluate every 500 steps. For ICL-D(LM) we set the max steps to 20000 and evaluate every 5000 steps. 

All the models in our work can be fit in a single NVIDIA RTX A6000. 

\section{Datasets}
\label{sec:appendix for datasets}
\subsection{In-context Learning Datasets}
\textbf{LAMA} Language Model Analysis is an entity prediction dataset for measuring the factual commonsense knowledge learned by LMs. We utilize the TREx-UHN portion of LAMA consisting of (subject, relation, object) triples from Wikidata. The LM is asked to predict the object entity given the subject entity and relation. We treat each relation as a single task \citep{Perez2021TrueFL}, resulting in 41 tasks and 16K examples in total. Following \cite{Petroni2019LanguageMA}, the object entity is predicted from a pre-defined vocabulary set of 21K words and therefore each task is 21K-way classification. We adopt mean precision at one and mean precision at ten as our evaluation metrics and report the average scores across tasks.
\\
\textbf{CrossFit}
CrossFit is a few-shot gym containing 160 diverse few-shot NLP tasks in a unified text-to-text format. We take 53 unique tasks which are closely related to real-life scenarios to evaluate our methods. We adopt Macro-F1 and Accuracy as evaluation metrics for classification tasks and non-classification tasks respectively.
\\ \\
\subsection{Language Modeling Datasets}
We leverage WikiText \citep{merity2016pointer} as the auxiliary dataset for LAMA and OpenWebText \citep{Gokaslan2019OpenWeb} for CrossFit. 
WikiText is extracted from the set of verified good and featured articles on Wikipedia, which contains much factual information. Therefore, we use the Wikitext-103-v1 as auxiliary language modeling data for experiments on LAMA. OpenWebText is another open-domain web text dataset collected by the scraper used to train GPT2. We sample 0.5GB of data from it as auxiliary language modeling data for experiments conducted on CrossFit. 

\section{Few-shot Settings}
We have four different few-shot settings in total. Setting 1: Meta-ICT on LAMA. Setting 2: Meta-ICT on CrossFit. Setting 3: Multitask-ICT on LAMA. Setting 4: Multitask-ICT on CrossFit.
Table \ref{table:tasks} shows the details for the tasks for different settings.
\label{sec:appendix for settings}
\begin{table*}[t]\small
\centering
\begin{tabularx}{\textwidth}{M{3cm}P{12cm}}
  \toprule
    \textbf{Setting} & \textbf{Tasks}\\
    \midrule
    Setting 1 train & P159, P1376, P27, P1412, P530, P108, P1303, P1001, P19, P30, P463, P413, P264, P740, P31, P39, P407, P279, P527, P495, P101, P17, P127, P20, P276, P36, P106, P176, P136, P937\\
    Setting 1 target & P140, P37, P138, P449, P361, P190\\
    \midrule
    Setting 2 train & superglue-rte, tweet\_eval-sentiment, discovery, glue-rte, superglue-wsc, glue-mrpc, tweet\_eval-stance\_hillary, tweet\_eval-offensive, emotion, hat- explain, glue-cola, sick, paws, ethos-sexual\_orientation, glue-qqp, tweet\_eval-emotion, sms\_spam, health\_fact, glue-mnli, imdb, ethos-disability, glue-wnli, sc-itail, trec-finegrained, yahoo\_answers\_topics, liar, glue-sst2, tweet\_eval-stance\_abortion, circa, tweet\_eval-stance\_climate, glue-qnli, tweet\_eval-emoji, ethos- directed\_vs\_generalized, ade\_corpus\_v2-classification, hate
    \_speech\_offensive, superglue-wic, google\_wellformed\_query, tweet\_eval-irony, ethos-gender, on-estop\_english, trec, rotten\_tomatoes, kilt\_fever\\
    Setting 2 target & tweet\_eval-stance\_feminist, ethos-national\_origin, tweet\_eval-hate, ag\_news, amazon\_polarity, hate\_speech18, poem\_sentiment, climate\_fever, medical\_questions\_pairs, tweet\_eval-stance\_atheism, superglue-cb, dbpedia\_14, wiki\_qa, emo, yelp\_polarity, ethos-religion, financial\_phrasebank, tab\_fact, anli, ethos-race \\
    \midrule
    Setting 3 train\&target & P19, P20, P279, P37, P413, P449, P47, P138, P364, P463, P101, P106, P527, P530, P176, P27, P407, P30, P178, P1376, P131, P1412, P108, P136, P17, P39, P264, P276, P937, P140, P1303, P127, P103, P190, P1001, P31, P495, P159, P36, P740, P361 \\
    \midrule
    Setting 4 train\&target &  ag\_news, amazon\_polarity, dbpedia\_14, emo, tweet\_eval-stance\_feminist, tweet\_eval-hate, superglue-cb, wiki\_qa, yelp\_polarity, quarel, glue-mrpc, qasc, commonsense\_qa, hate\_speech18, superglue-copa, sciq, glue-sst2, ethos-religion\\
  \bottomrule
\end{tabularx}
\caption{\label{table:tasks} Tasks for different settings}
\end{table*}

\section{In-context Learning}
\label{sec:appendix for channel}
There are two types of in-context learning depending on the order of input texts and labels.
\\
\textbf{Direct In-context Learning} 
In direct in-context learning, the model is given a concatenation of $x_1, y_1, ..., x_k, y_k, x_{k+1}$ as input and predicts $y_{k+1}$. 
More formally, the model computes $argmax_{c\in C}P(c|x_1, y_1, ..., x_k, y_k, x_{k+1})$ to do inference, where $C$ is a set of label candidates. 
\\
\textbf{Channel In-context Learning} 
There is another noisy channel variant of in-context learning objective called channel in-context learning \citep{min-etal-2022-metaicl}, in which the input $x_i$ and $y_i$ is flipped.  Channel in-context learning converts $P(y|x)$ into
$\frac{P(x|y)P(y)}{P(x)} \propto P(x|y)P(y)$. 
 We let $P(y)$ be constant and aim at modeling $P(x|y)$.  During inference time, the model computes $argmax_{c\in C}P(x_{k+1}|y_1, x_1, ..., y_k, x_k, c)$ where $C$ is a set of candidates. Then the model returns the label with the maximum conditional probability as the prediction. 
\end{document}